\documentclass[10pt,conference]{IEEEtran}
\makeatletter

\IEEEoverridecommandlockouts                              

\usepackage{amssymb, amsmath}
\usepackage{graphicx}
\usepackage[vlined,ruled]{algorithm2e}
\usepackage{color}
\usepackage{multirow}
\usepackage{float}
\usepackage{verbatim}
\usepackage{tabularx}
\usepackage{url}

\usepackage{subfig}
\usepackage{xspace}
\usepackage{balance}
\usepackage{makecell}
\usepackage{tabularx}

\newcommand{\system}{TCAR\xspace}
\newcommand{\framenet}{FrameNet\xspace}

\begin{document}

\title{Enabling human-like task identification \\from natural conversation}

\author{\IEEEauthorblockN{Pradip Pramanick, Chayan Sarkar, Balamuralidhar P, Ajay Kattepur, Indrajit Bhattacharya and Arpan Pal}
	\IEEEauthorblockA{TCS Research \& Innovation, India}
}

\maketitle
\thispagestyle{empty}
\pagestyle{empty}

\begin{abstract}
A robot as a coworker or a cohabitant is becoming mainstream day-by-day with the development of low-cost sophisticated hardware. However, an accompanying software stack that can aid the usability of the robotic hardware remains the bottleneck of the process, especially if the robot is not dedicated to a single job. Programming a multi-purpose robot requires an on the fly mission scheduling capability that involves task identification and plan generation. The problem dimension increases if the robot accepts tasks from a human in natural language. Though recent advances in NLP and planner development can solve a variety of complex problems, their amalgamation for a dynamic robotic task handler is used in a limited scope. Specifically, the problem of formulating a planning problem from natural language instructions is not studied in details. In this work, we provide a non-trivial method to combine an NLP engine and a planner such that a robot can successfully identify tasks and all the relevant parameters and generate an accurate plan for the task. Additionally, some mechanism is required to resolve the ambiguity or missing pieces of information in natural language instruction. Thus, we also develop a dialogue strategy that aims to gather additional information with minimal question-answer iterations and only when it is necessary. This work makes a significant stride towards enabling a human-like task understanding capability in a robot.
\end{abstract}

\section{INTRODUCTION}
Recent advancements in robotics see rapid inroads of robots into our daily surroundings as helper~\cite{pramanick2018defatigue}, companion~\cite{vu2015companion} or coworker~\cite{hinds2004whose}. Being able to execute tasks that are conveyed in natural language, is the most sought after feature in modern robotics. Recent advancements in natural language processing (NLP) has enabled robots to interact with human cohabitants and collaborators in natural language. Yet the ambiguity present in natural language makes it very difficult for a robot to fully interpret the task goals and perform the task conforming to the human intention. Human beings generally converse in short sentences, often with many implicit assumptions about the task context. To overcome this limitation, Matuszek \textit{et al.}~\cite{matuszek2013learning} proposed a restricted natural language based interaction with the robot. This not only helps to mitigate the possible ambiguity in the interaction, but it also eases the process of suitable plan generation to execute the task. However, in a multi-purpose robot, the set of capabilities can be large and programming the robot for each and every task is cumbersome. Moreover, a study by Thomason \textit{et al.}~\cite{thomason2015learning} suggests that restricted natural language limits the usability and acceptability of the robot, especially in daily surroundings like home, office, hospital, restaurants, etc.
\begin{figure}
	\centering
	\includegraphics[width=0.9\linewidth]{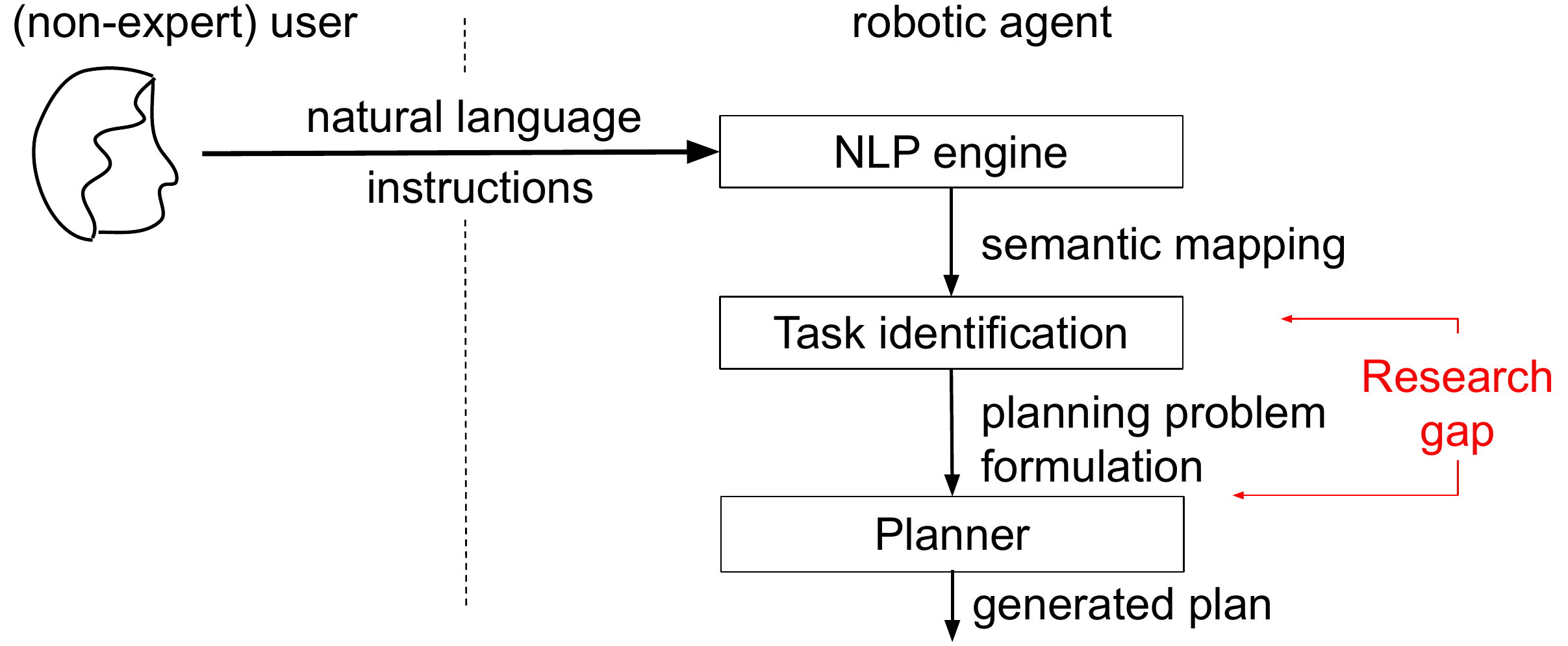}
	\caption{A generic high-level system to generate task plan from natural language instructions.}
	\label{fig:building_blocks}
	\vspace{-0.3cm}
\end{figure}

After getting instructions in natural language, executing the desired tasks involves several internal processing steps by the robot. Fig.~\ref{fig:building_blocks} depicts a high-level building block for such a system. Clearly, an NLP engine is a necessary part of this system, but not sufficient. There are multiple issues that need to be tackled. Firstly, a general-purpose NLP engine only provides syntactic details of natural language instruction. They do not accompany mechanism that can classify a sentence as a task for the robot, type of task, and the parameter set hidden within the sentence. This requires a domain-specific knowledge of the world where the robot is operating and the capability set of the robot. Secondly, the ambiguity in any natural conversation poses a challenge in identifying human intention using one-way interaction. Thus, a bidirectional conversation is a necessity. Though there exists a number of work on conversational systems~\cite{agarwal2017continuous,lam2017social,patidar2018automatic,radlinski2017theoretical}, systems that assign tasks to robots are rare. 

In this work, we propose ``\textbf{T}ask \textbf{C}onversational \textbf{A}gent for \textbf{R}obots (TCAR)'' that contains a task classifier and a dialogue engine along with essential NLP toolset and a planner. The task classifier is trained with a known set of tasks that are mapped to the high-level capabilities of the robots. Each task is associated with a template that consists of pre- and post-conditions. Using a temporal grounding graph model, Paul \textit{et al.}~\cite{paul2018temporal} proposed a world modeling mechanism. TCAR accompanies a similar knowledge-base (KB) along with the task templates. Given a natural language instruction, TCAR reduces this as the problem of encoding a set of grounded logical propositions that represent the expected initial and the final state of the world for the task and then logically reasons with the symbolic state of the world. Finally, it formulates a planning problem and uses a planner to solve the problem, which provides an output as the required sequence of primitive actions.


Major contributions of this work are summarized in the following.
\begin{itemize}
	\item We develop a context-aware planning problem formulator for robots that takes task instruction in unrestricted natural language and converts it into a planner consumable problem.
	
	\item The system can accurately identify the task(s) and the associated parameters from the natural language instruction using a classification tool-chain and pinpoint the ambiguity in the instruction if any.
	
	\item Our dialogue strategy can help to resolve ambiguity using guided conversation. Also, it can quickly identify the tasks that are beyond the capacity of the robot.
\end{itemize}

\section{RELATED WORK}
There exists a plethora of work on natural language understanding. Here, we discuss the most relevant works that focus on understanding instructions given to a robot and highlight the challenges in interpreting abstract, ambiguous, and often incomplete natural language instructions.

Some early works by Lauria \textit{et al.}~\cite{lauria2002converting} and Kollar \textit{et al.}~\cite{kollar2010toward} have attempted to execute navigational instructions given in natural language, where tasks are translated to a first-order logic form and a semantic tree structure to extract action procedures and the corresponding arguments.  In more recent works (by Chen \textit{et al.}~\cite{chen2011learning} and Matuszek \textit{et al.}~\cite{matuszek2013learning}), supervised learning approaches are used to learn semantic parsers that map route instructions to primitive actions and control expressions. Though they provide interesting directions in understanding natural language instructions, they overlook many challenges by restricting to only navigation actions.



Semantic parsers that are based on rich lexical resources such as \framenet\cite{baker1998berkeley} and Propbank~\cite{palmer2005proposition} are proven to be useful in general-purpose natural language understanding problems. \framenet, which focuses on the meaning of common verbs, has been used to parse natural language instructions to predicate-argument structures in~\cite{scheutz2007first,thomas2012roboframenet}. However, these parsers alone cannot completely remove the ambiguities before executing the instructions physically by a robot, because they only take the linguistic information as input and do not consider the context set by the details of the world model. Bastianelli \textit{et al.}~\cite{bastianelli2016discriminative} have tried to resolve some of these ambiguities by developing a semantic parser for \framenet, which predicts task using both linguistic information and the perceived world model of the robot. However, this requires a complete and consistent semantic map of the environment, which is difficult to obtain in a dynamic environment. In real scenarios, robots are equipped with very low-level action procedures and natural human commands often contain very high-level task goals, which makes it very difficult to learn such a mapping. It is also difficult to anticipate all the possible sequences of the primitive actions for a task in different contexts. Therefore, planning has been widely used to determine the required primitive action sequences. However, generating input for a planner from an unstructured and ambiguous natural language is still an active area of research.



Thomas and Jenkins~\cite{thomas2012roboframenet} proposed Robo\framenet, a framework for parsing human instructions to semantic frames of \framenet and then generate the primitive action sequences. They have used a handcrafted lexical resource to predict the task and a generic dependency parser to extract the action arguments, which limits the language understanding capabilities. Also, finding the action sequence using a random search is inefficient and not extendable to real scenarios. 

Bollini \textit{et al.}~\cite{bollini2013interpreting} proposed a system to learn appropriate baking primitives from recipes available on the internet, using a reward function. Such an end-to-end approach to ground instructions with context-sensitive planning capabilities has been explored further in recent works. Antunes \textit{et al.}~\cite{antunes2016human} proposed architecture of probabilistic planning to try out different actions on sub-task failure and re-ordering of sub-goals towards a  successful plan generation for a complex task. Similarly, Misra \textit{et al.}~\cite{misra2016tell} also proposed an action sequence grounding method by adding, removing and replacing actions from learned samples in similar contexts. Such approaches to learning primitive action sequences for a given task do not generalize well in new domains and robots with different capabilities. Also, substantial effort is required to create new datasets for training the robot to work in a new domain.


Our proposed system differs from the end-to-end and supervised learning approaches to learn primitive action sequences from natural language instructions, in many ways. We use a general-purpose semantic parser to extract task and arguments from natural instructions. Understanding instructions that contain novel verbs are often handled by word similarity measures using WordNet~\cite{chen2013handling,lu2017interpreting} or by using environment-specific training data~\cite{misra2015environment} for disambiguation. In contrast, we propose a dialogue strategy to understand novel and ambiguous instructions, which leads to better instruction interpretation and shorter interactions. Also, the system generates input for domain-independent, off-the-shelf planners in PDDL formal language. This lessens the effort to re-use our system in a new domain. Though Lu \textit{et al.}~\cite{lu2017integrating} and Munawar \textit{et al.}~\cite{Munawar2018} proposed approaches to convert the parsed semantic information to a formal language for planners, we extend this by a complete framework that uses human-robot dialogues and context-sensitive task planning, leading to highly accurate conversion of natural language instructions to planned sequences of primitive actions. Moreover, we also tackle the problem of handling compound instructions containing multiple tasks.

\section{SYSTEM OVERVIEW}

\begin{figure*}[ht]
	\centering
	\includegraphics[width=\linewidth]{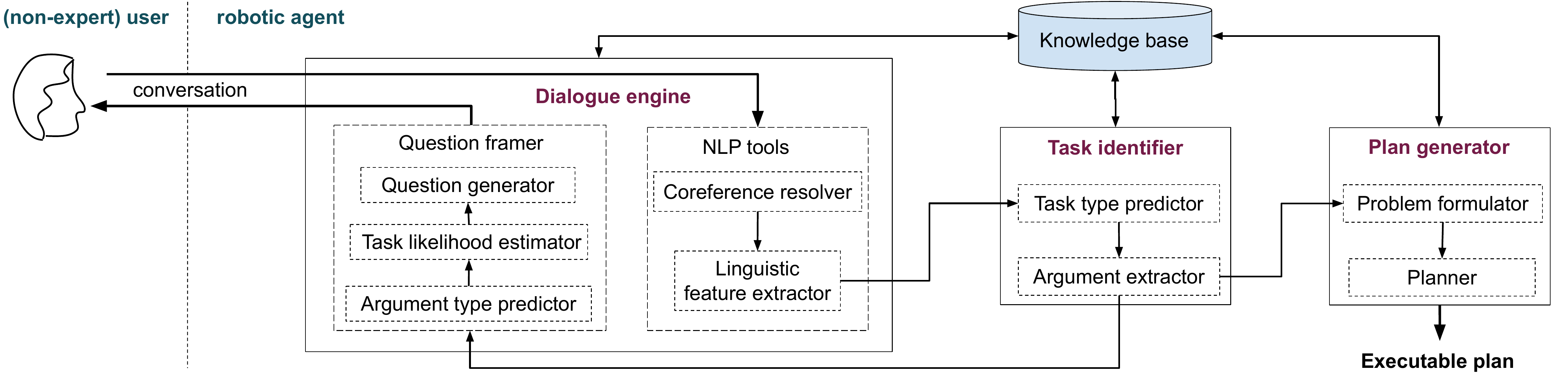}
	\caption{Overview of \textbf{T}ask \textbf{C}onversational \textbf{A}gent for \textbf{R}obots (\system) with major building blocks.}
	\label{fig:robo_brain}
\end{figure*}
In this section, we provide the design philosophy of \system before describing the finer details in the subsequent sections. \system consists of four main parts as shown in Fig.~\ref{fig:robo_brain} -- (i) a \textit{dialogue engine} that handles bidirectional interaction between a robot and an user, (ii) a \textit{task identifier} that identifies the intended task and the relevant parameters from the interaction, (iii) a \textit{plan generator} that ensures a valid plan is generated for a given task, and (iv) a \textit{knowledge-base} that contains the world model where the robot is operating and the task templates for plan generation.


\subsubsection{Dialogue engine} It consists of a set of generic \textit{NLP toolset} that extract the features from the utterance text received from the user and a \textit{question framer} that formulate relevant questions (only if it is necessary) for the user to resolve ambiguity in task understanding. For example, if the user says, \textit{``Take the book from the table''}, the NLP tools process it to provide the following output consisting of grammatical features. 

\textit{\small (Take, VB, root), (the, DT, det), (book, NN, dobj), (from, IN, prep), (the, DT, det), (table, NN, pobj)}.

\subsubsection{Task identifier} To remove ambiguity and to understand the semantic information provided by the NLP toolset, a common vocabulary has to be agreed upon by the robot and the human. Even though the human provides the instruction in natural language, the same can be converted to an intermediate representation (with uncertainty) that the robot can store and process. We take help from the \textit{Frame Semantics} theory, specifically building on the formalization of the \textit{FrameNet} project~\cite{baker1998berkeley} to achieve this task modeling. Frame semantics models an event in the physical world as a \textit{frame}, which can completely describe the event using it's participating entities, called \textit{frame elements}. For example, an event of taking an object from a location is modeled with a \textit{taking} frame. To describe the event, the frame elements \textit{theme} and \textit{source} are used, where \textit{theme} represents the object affected by the event and \textit{source} represents a location where the \textit{theme} is present. Thus, when the output of the NLP tool is processed by the task identifier, it produces the following output. 

\textit{\small [Take]\textsubscript{taking}  [the book]\textsubscript{theme}  [from the table]\textsubscript{source}}.

\noindent We developed a classifier to identify the frame and the frame elements. If the classifier fails to substantiate with sufficient confidence, the user is asked relevant questions to resolve the ambiguity and missing piece(s) of information.

\subsubsection{Plan generator} One-to-one mapping is often not possible between a human intended task and the primitive actions supported by the robot, because a high-level task goal may require performing a sequence of sub-tasks. To do the same, task planning techniques have been used to compute the required sequence. To enable task planning, the state of the world is exposed to the robot in terms of grounded \textit{fluents}, which are logical predicates that can have variables as arguments. A task starts from an initial state of the world and leads to a different state of the world, namely the goal state.

The Planning Domain Definition Language (PDDL)~\cite{mcdermott1998pddl}, widely used in classical planning, provides a method for encoding these initial and goal state of a world in a formal language. However, to be able to execute the planned action sequence, each robot action must also be mapped to a PDDL state-transition operator. In PDDL terminology, the definitions of the set of fluents, argument variables, and the state-transition operators is called a \textit{domain}, and the set of grounded initial and goal condition is called a \textit{problem}. A symbolic planner can solve the planning problem in a given domain, thereby finding the sequence of robot actions that satisfy the human intended task goal.

\subsubsection{Knowledge-base} It stores the world model and task templates. World model contains symbols that represent the current state of the world where the robot is operating. Similar to the model of perceptual grounding of world state proposed by Paul \textit{et al.}~\cite{paul2018temporal}, \system frequently updates the world model as the robot perceives changes in the environment. The KB also provides the task context that is taken into consideration while generating the planning problem for the task planner. We also update the KB after actions are executed by the robot. The update adds new and non-contradicting fluents to the KB and replaces the old ones with the new fluent in case of contradiction. On the other hand, task templates store the possible pre-condition and post-condition that are used to generate the planning problem for the task, which is compliant with the frame semantics. If some essential arguments are missing for a given task instruction, the question framer initiates a dialogue with the user to gather further information.

\section{PILLARS OF \system}\label{pillars}
A high level architecture of \system is shown in Fig.~\ref{fig:robo_brain}. Though many components of \system are borrowed from state-of-the-art, we develop three indigenous components that help \system achieve its goal and also maintain its novelty.

\subsection{Task and argument identification}
Given a piece of syntactically tagged text (tokens), the role of task identifier is to identify the intended task(s) and their corresponding arguments. This can be formulated as a classification problem to classify a word or a phrase to a task type or an argument. We use ``Conditional Random Field'' (CRF) models for this purpose. CRF is a discriminative model for text sequence labeling and is proved to be useful in similar problems~\cite{kollar2013learning,misra2015environment}. The parsing is done in two sequential stages -- task type prediction and argument extraction. The task type prediction stage predicts possible sequences of frames in a given text. Then the argument extraction stage predicts sequences of frame elements in the text, given the prediction of possible frames in the previous stage. The predicted sequences are labeled by predicting IOB tags for each token of the text that denotes whether the token is inside (I), outside (O) or at the beginning (B) of the label.

For the task type identification stage, the training data is given as, \[[s_j=[w_i,tt_i]_{i=1}^M]_{j=1}^N,\] where a sentence $s_j$ is given by a $M\times2$ matrix containing the words ($w_i$) and their corresponding IOB tag of the task type ($tt_i$) and $N$ is the number of such sentences in the training data. For the argument extraction phase, the training data is given as, \[[s_j=[w_i,T_i,at_i]_{i=1}^M]_{j=1}^N,\] where $T_i$ is the predicted task type associated with the word $w_i$ and $at_i$ is the IOB tag of the argument type. The CRF model for task type identification defines a conditional probability distribution,
\[P(tt_{1:M}|w_{1:M})= \alpha \exp(\sum\limits_{M} \sum\limits_K W_k \phi_k(tt_{i-1},tt_i,w_i)),\]
where $\phi_k$ is the $k^{th}$ component of the feature function, k is the number of features, $W_k$ is the weight of the $k^{th}$ feature and $\alpha$ is a normalization factor. The weights are learned from the training data using a gradient descent optimization.

For the argument extraction stage, the CRF defines the following conditional probability distribution.

\[P(at_{1:M}|w_{1:M})= \alpha \exp(\sum\limits_{M} \sum\limits_K W_k \phi_k(at_{i-1},at_i,w_i,T_i)).\]

We have used lexical and grammatical features to predict both the task and argument types. Lexical features include the word itself, its lemma and the words of the left and right context. Grammatical features include parts of speech and syntactic dependency of the word and the context words. We extract the features using a generic NLP library, Spacy\footnote{https://spacy.io/}.


\subsection{Task understanding by conversation}\label{Dialouge-similarity}
Even in the presence of an accurate task identifier, ambiguity in natural language instruction may lead to identification failure or misprediction, especially when it comes from a non-expert user. Existing task identification models are generally trained with the verbs that are present in the instruction and the linguistic features around the verbs~\cite{bastianelli2016discriminative,liu2018generating,lu2017integrating,misra2016tell,thomas2012roboframenet}. However, a non-expert user may use verbs that are unseen for the model or use an ambiguous usage of the verb not present in the training data. In such cases, the task identification model may mispredict the task type or may not be able to predict with high confidence. As the task type identification is a fundamental stage in the processing pipeline, its failures and errors need to be resolved to be able to execute the instruction. 

Traditionally, in these scenarios, the robot engages with the human in a conversation to determine the meaning of the novel instruction or the correct task type in case of misprediction~\cite{thomason2015learning}. For example, if the robot can not predict the task type, it can ask the human for the same. However, a non-expert user may not be aware of the terminologies used by the robot. Thus, a non-expert may not be able to give correct answers to direct questions such as, \textit{``what type of task is this?''} simply because he/she doesn't know or remember what are the task types the robot knows and by what convention they are categorized. So, the robot must inform the non-expert user of its knowledge of task types. To do this the robot can ask suggestive questions, such as, \textit{``Is this task similar to (suggestion)?''}  In this case, the human can give a direct yes/no answer, which is more likely to be correct. 

This scheme gives rise to two new problems. Firstly, the number of task types known by a robot is usually not very small (11 types of task in our system). If the robot suggests them one by one, it degrades the user experience badly. So, the robot should only ask about the most probable task types. Secondly, a non-expert may not be unable to understand the suggestion itself as he/she is unaware of the convention of defining the task type. In this case, the user may ask to clarify the meaning of the task. We propose a dialogue strategy to handle these two problems.

For a given instruction, decoding the task can be jointly characterized by both the verb and the nouns phrases that act as the arguments of the verb. We exploit these task-argument relationships present in a dataset (the same dataset used to train the parser) to estimate the likelihood of a known task being conveyed. Formally, given a sentence $S$ and a set of all possible task types, $T=\{T1,T2....Tn\}$, we estimate a n-tuple, $T'$, such that each element $T'_i$ denotes a task type from $T$ and the sequence of $T'$ is given by the likelihood of $T'_i$ being the true task type for the sentence, i.e., $P(T_i|S)$. We estimate $T'$ using the following procedure. Firstly, we use an argument type predictor model, which is used to find the possible argument types present in the instruction. The argument type predictor model is also realized as a CRF.
\[P(at'_{1:M}|w_{1:M})= \alpha \exp(\sum\limits_{M} \sum\limits_K W_k \phi_k(at'_{i-1},at'_i,w_i)).\]
This model is different from the argument extraction model by the fact that this model predicts the IOB tags of the argument type ($at'_i$) for each word ($w_i$) without considering the task types for the sentence. Secondly, the predicted $at'_i$ are converted to a set of argument types in the sentence, given by $AT_P$. We define another set $AT_D$ as the set of argument types of a task type, present in an instruction in the training dataset. The number of instances of task type $T_i$ that satisfies $AT_P \subset AT_D$ is counted for all the instructions in the training dataset. This generates an n-tuple, where the elements are from the set $T$, ordered by the corresponding counts. The counts are normalized to convert it into a probability distribution, and then the n-tuple is sorted by the probabilities, which finally gives $T'$. After asking about all the task types in $T'$, the dialogue strategy determines that the robot is unable to perform the task and an expert's intervention is required.

To answer the user's questions about the definitions of the task types, our model uses a question template for every task type. The template contains slots which are filled by the predictions of the argument type predictor model. For example, if a non-expert is asked to clarify the meaning of placing task, the template, \textit{``Do you want me to put [\textsubscript{theme}] in [\textsubscript{goal}]?''}, is used. The dialogue for the same example is shown below.
\begin{description}
	\small
	\item    H1: add some water to the bowl 
	\item    R1: Is this task similar to placing ? 
	\item    H2: I didn't understand
	\item    R2: Do you want me to put some water in the bowl ? 
	\item    H3: Yes 
	\item    R3: Got it. 
\end{description}

We use similar templates to ask questions when missing arguments are to be identified. For example, if the source is missing from an instruction of a taking task, question template, \textit{``From where do I take it?''}, is used. A high-level task specified by a non-expert can also be a composition of the known tasks. We have also enabled \system with the capability to interact with the user to extract the sequence of known tasks. In this dialogue scenario, the robot asks the user to say the steps to perform the high-level task. Then the response is treated as a single instruction containing multiple serialized tasks, which \system takes as input to generate the plan.

\subsection{Planning Problem Formulation}
A task given as a natural language instruction expresses a specific goal to be fulfilled. Also, the task can be assumed to be given at a hypothetical world state, which is the initial state for the task. We store the templates of the initial and the goal states for each of task types in the KB. Table~\ref{table:templates} shows some examples of such templates. From such a template, we generate the grounded initial and the goal states of the planning problem by first grounding the variables of the fluents using the arguments extracted from the instruction and then by validating using the world model. We create the templates from the definition of the frame that models the task, also considering the predicates of the PDDL domain. The effort for template creation is manageable because templates are created once and there is a small number of task types to be considered.
\begin{table}
	\centering
	\caption{Example of templates for generating initial and goal state in PDDL.}
	\label{table:templates}
	\begin{tabular}{|p{1.4cm}|p{2.9cm}|p{2.9cm}|}
		\hline
		\textbf{Task} & \textbf{Initial state template} & \textbf{Goal state template} \\ \hline
		Bringing & hasobject(source, theme) & hasobject(goal, theme) \\ \hline
		Change-state & near-device(device) & current-state(device, state) \\ \hline
		Motion & $\emptyset$ & at-robot(goal) \\ \hline
		Placing & holds(theme)  & hasobject(goal, theme) \\ \hline
		Taking & hasobject(source, theme) & holds(theme) \\ \hline
	\end{tabular}
\end{table}

If a robot works autonomously, it continuously accepts new tasks from a human instructor and performs them. While generating a planning problem for a new task, we use a world model to validate the assumed initial state. We get the fluents that encodes the state of the world as modeled by the planner using the post-conditions of the state transition operators in the computed plan. We use the \textit{Metric FF planner}~\cite{hoffmann2001ff} to get this resultant state. This planner-modeled world state is valid only if a closed world assumption holds true. Our model allows relaxation of this assumption to some extent by fusing the information coming from the robot's perception systems. We assume that this information is also encoded as fluents to support reasoning in the knowledge base. To make this knowledge consistent, the fluents predicted by the planner that contradicts with the fluents from the perception sub-systems, are replaced by the later. The full procedure for generating the planning problem is shown in Algorithm~\ref{Algo:Problem Formulation}. For the instructions that contain multiple task goals to be fulfilled, we assume the tasks are serialized. We handle such compound instructions by preserving the world state context across all the tasks in the instruction. The stored template mechanism also eases the integration of new tasks or to use our system in a different domain. A new task can be added by introducing a new template along with the predicates and state transition operators in PDDL domain.
{
	\begin{algorithm}[h]
		\small
		\LinesNumbered
		\SetAlgoLined \DontPrintSemicolon
		\SetKwInOut{Input}{Input}\SetKwInOut{Output}{Output}
		\SetKwInput{Initialize}{Initialization}
		\Input{Goal state template: T\textsubscript{G}, Initial state template: T\textsubscript{I}, World state: W, Parsed arguments: P }
		
		\Initialize{init\_state = $\emptyset$, goal\_state = $\emptyset$
		}
		\SetKwProg{myalg}{Algorithm}
		\myalg{\algo}{
			\While {T\textsubscript{I} $\neq \emptyset$}{
				\text{Pop template\_atom from T\textsubscript{I}}\;
				\text{grounded\_atom=Ground(template\_atom, P)}\;
				\While{W $\neq$ $\emptyset$}{
					\text{Pop fluent from W}\;
					\If{fluent contradicts with grounded\_atom}{
						\text{add fluent to init\_state}\;
					}
					\Else{
						\text{add grounded\_atom to init\_state}\;
					}
				}
			}
			\While{T\textsubscript{G} $\neq \emptyset$}{
				\text{Pop template\_atom from T\textsubscript{G}}\;
				\text{grounded\_atom=Ground(template\_atom, P)}\;
				\text{add grounded\_atom to goal\_state}\;
			}
			\Output{init\_state, goal\_state}   
			
		}
		\caption{\small Generation of PDDL planning problem.}
		\label{Algo:Problem Formulation}
	\end{algorithm}
}

\section{EVALUATION}
To test and validate our system, we have used the HuRIC corpus~\cite{bastianelli2014huric} to train our language understanding models introduced in Section~\ref{pillars}. The performance of the task and argument identification models on the test data (80:20 train-test split) of HuRIC is shown in Table~\ref{tab:test-performance}.
\begin{table}
	\centering
	\caption{Accuracy of the language understanding modules on HuRIC dataset.}
	\label{tab:test-performance}
	\begin{tabular}{|c|c|c|c|}
		\hline
		\textbf{Model} & \textbf{Precision} & \textbf{Recall} & \textbf{F1} \\ \hline
		Task type predictor & 92 &91 &91 \\ 
		Argument extractor & 93 & 94 & 93 \\
		Argument type predictor & 83 & 86 & 83 \\ \hline
	\end{tabular}
\end{table}
To evaluate our system in terms of plan generation, we have used a natural language instruction data-set from a well-known competition Rockin@Home\footnote{http://rockinrobotchallenge.eu/home.php}. This competition aims to develop robust robotic systems that work in a house environment as a helper to the elderly. The dataset is divided into four groups based on the data from similar competitions. Each group contains a set of audio files with their transcriptions and annotations using FrameNet. We take the transcriptions as input to our system and evaluate its task understanding and planning capabilities. To evaluate the performance of our proposed dialogue strategy for task disambiguation, we have used the VEIL dataset described in~\cite{misra2016tell}. The VEIL dataset contains human-provided instructions to perform different tasks, also in a domestic service robotics scenario. The instructions in VEIL are more natural, ambiguous and contains many novel verbs that our task identification model is not trained with.

\subsection{Performance of the task identifier}
While developing this end-to-end system, combining various processing stages, we have identified several issues and proposed solutions for them. In the following, we show how our system performs while solving these issues comparing against a baseline method. Table~\ref{table:models} shows a brief description of the different systems with various modules plugged-in to solve the issues discussed earlier.
\begin{table}
	\centering
	\caption{Different methods used for task understanding and plan generation.}
	\label{table:models}
	\begin{tabular}{|p{2.1cm}|p{3.2cm}|p{1.9cm}|}
		\hline
		\textbf{System} & \textbf{Instruction understanding} & \textbf{Plan generation} \\
		\hline
		Baseline & Using semantic parser alone & Static templates \\
		\hline
		Interactive task\newline understanding (\system-I\textsubscript{d}P\textsubscript{0}) & Using dialogue for missing information along with the parser & Static templates \\
		\hline
		Interactive task\newline understanding and contextual\newline planning (\system-I\textsubscript{d}P\textsubscript{c}) & Using dialogue for missing information along with the parser & Templates updated by\newline world model \\
		\hline
		Complete plan\newline generation model & Using a co-reference resolver along with the parser, dialogue for missing information & Templates updated by\newline world model\\ \hline
	\end{tabular}
\end{table}
By comparing with the annotations in Rockin@Home dataset, we found that our frame semantic parser can correctly identify 420 out of 439 (95.7\%) tasks present in total 393 instructions (shown in Table~\ref{table:task-identification-accuracy}). Clearly, the parser is very accurate in predicting the task types from natural language instructions. However, we show in Table~\ref{tab:result-rockin} that even the presence of such a highly accurate parser, the performance of the baseline system degrades considerably in plan generation. The baseline system generates plans for 191 tasks, which is only 43.5\% of the total tasks. This is because in many of the instructions, one or more arguments are missing and the baseline system doesn't use dialogues to get the missing information. Also, because of static templates, planning problems are not generated for the instructions that contain multiple tasks with conflicting goal states. 
\begin{table}
	\centering
	\caption{Performance of \system for task identification on the Rockin@Home dataset.}
	\label{table:task-identification-accuracy}
	\begin{tabular}{|p{1.5cm}|p{2.0cm}|p{1.5cm}|p{1.6cm}|}   
		\hline
		\textbf{Group} & \textbf{\# of instructions} & \textbf{\# of tasks} & \textbf{\# of correct predictions}  \\
		\hline
		Robocup &144 & 163 & 153 (93.8\%)  \\ 
		\hline
		Rockin1 &115 & 134 & 129 (96.3\%) \\
		\hline
		Rockin2 &114 & 120 & 117 (97.5\%) \\
		\hline
		Rockin2014 &20 & 22 & 21 (95.5\%) \\
		\hline
	\end{tabular}
\end{table}
\begin{table}
	\centering
	\caption{Performance of \system for plan generation on the Rockin@Home dataset.}
	\label{tab:result-rockin}
	\begin{tabular}{|c|c|c|}
		\hline
		\textbf{System} & \textbf{Group} & \textbf{Plan generated} \\
		\hline
		Baseline & Robocup  & 95 (58.3\%) \\
		& Rockin1  & 53 (39.5\%)  \\\
		& Rockin2  & 35 (29.1 \%) \\
		& Rockin2014 & 8 (36.3\%) \\
		\hline
		\system-I\textsubscript{d}P\textsubscript{0} & Robocup & 138 (84.6\%) \\
		& Rockin1 & 110 (82.1\%) \\
		& Rockin2 & 98 (81.6\%) \\
		& Rockin2014 & 19 (86.3\%) \\
		\hline
		\system-I\textsubscript{d}P\textsubscript{c} & Robocup & 148 (90.7\%) \\
		& Rockin1 & 120 (89.5\%)  \\
		& Rockin2 & 103 (85.8\%)  \\
		& Rockin2014 & 21 (95.4\%) \\
		\hline
		Complete plan generation model & Robocup & 152 (93.2 \%) \\
		& Rockin1 & 122 (91.0\%) \\
		& Rockin2 & 105 (87.5\%) \\
		& Rockin2014 & 21(95.4\%) \\
		\hline
	\end{tabular}
\end{table}

By adding the dialogue module to get the missing arguments (\system-I\textsubscript{d}P\textsubscript{0}), the performance improves by a high degree, as shown in Table~\ref{tab:result-rockin}. \system-I\textsubscript{d}P\textsubscript{0} generates a total of 333 plans, which is 83.1\% of the the total tasks. To be able to evaluate such a large number of instructions, we have used a simulated human participant. The simulated participant gives the correct answer to the question about a missing argument if that argument is not present in the instruction; otherwise, it does not provide an answer. This dialogue solves the problem of incomplete instructions, but complex instructions that require context-sensitive planning, can not be handled by the static templates. This is improved further by the (\system-I\textsubscript{d}P\textsubscript{c}) model, which generates plans for 392 tasks or 89.3\% of the total tasks. Even though the (\system-I\textsubscript{d}P\textsubscript{c}) model generates plans for many instructions that contain dependent sub-tasks with conflicting goal conditions, it is unable to do so for some instructions where \textit{Anaphora} is used to refer entities, e.g., \textit{Take the pen and bring it to me}. 

We have used a state of the art co-reference resolver\footnote{https://github.com/huggingface/neuralcoref} that takes a text and returns it with the pronouns replaced by the nouns they are referring to. This leads to successful plan generation for 400 tasks or 91.1\% of the total tasks. This matches closely with the percentage of tasks correctly understood (95.7\%). We have investigated the reasons for the tasks that are predicted correctly but valid plans are not generated. This is mainly because the simulated human does not provide the arguments that are already present in the instruction. Also in some scenarios, a planning failure of a task, leads to failures of the dependent tasks in the same instruction, because of incorrectly assumed context.

\subsection{Evaluation of the dialogue strategy}
\system uses a dialogue strategy to generate plans for instructions that are incorrectly parsed, either because it contains a novel verb or the instruction is ambiguous. In both cases, one mandatory question is asked to verify whether the original prediction (with low confidence) is correct or not. If the original prediction is correct, then the system proceeds with plan generation; otherwise, it starts to ask questions about the similarity of the given task with the known tasks. The strategy described in Section~\ref{Dialouge-similarity} provides a sequence of questions so that the correct answer can be found by asking a minimal number of questions. We have evaluated this dialogue strategy against a baseline strategy that uses WordNet~\cite{miller1995wordnet}. The baseline strategy is motivated by the fact that WordNet has been used to find semantically similar tasks~\cite{chen2013handling,lu2017interpreting}. This baseline computes the similarity between the verbs that are most commonly used (based on training dataset) to specify a task. Then it provides the list of questions to be asked by ranking using the similarity score given by WordNet.
\begin{table}
	\centering
	\caption{Instructions containing a novel verb (in boldface) and the most similar task type.} 
	\label{tab:novel-verbs}
	\begin{tabular}{|l|l|}
		\hline
		\textbf{Instruction} & \textbf{Task type}\\
		\hline
		\textbf{add} some water to the bowl & Placing  \\ \hline
		\textbf{gather} all the cups & Bringing \\ \hline
		\textbf{dump} the bowl into the trash & Placing \\ \hline
		\textbf{drop} it in trash can  & Placing \\ \hline
		\textbf{grasp} the book & Taking \\ \hline
		\textbf{set} some pillows on the couch too & Bringing \\ \hline
		\textbf{pour} the contents of the pot into a bowl & Placing \\ \hline
		\textbf{throw} the bottle in the trash & Placing \\ \hline
		\textbf{collect} the cups from the table & Taking \\ \hline
		\textbf{release} the bag & Placing \\ \hline
	\end{tabular}
\end{table}
\begin{table}
	\centering
	\caption{Examples of ambiguous instructions that results in incorrect initial prediction compared to the intended task type. Through dialogue, finally the intended task type is retrieved  (Fig.~\ref{fig:ambigous-verbs}).} 
	\label{tab:ambigous-verbs}
	\begin{tabular}{|p{3.3cm}|p{1.7cm}|p{2.1cm}|}
		\hline
		\textbf{Instruction} & \textbf{Intended task} & \textbf{Initial prediction} \\ \hline
		move the remote near tv	& Bringing	& Motion \\ \hline
		turn to the right & Motion & Change-state \\ \hline
		put on the tv & Change-state & Placing \\ \hline
		keep the same pace as he has &Following	&Taking \\ \hline
		go to the kitchen with him &Following &Motion \\ \hline
		take the tray to the bedroom &Bringing &Taking \\ \hline
	\end{tabular}
\end{table}

\begin{figure}[t]
	\centering
	\includegraphics[width=\linewidth]{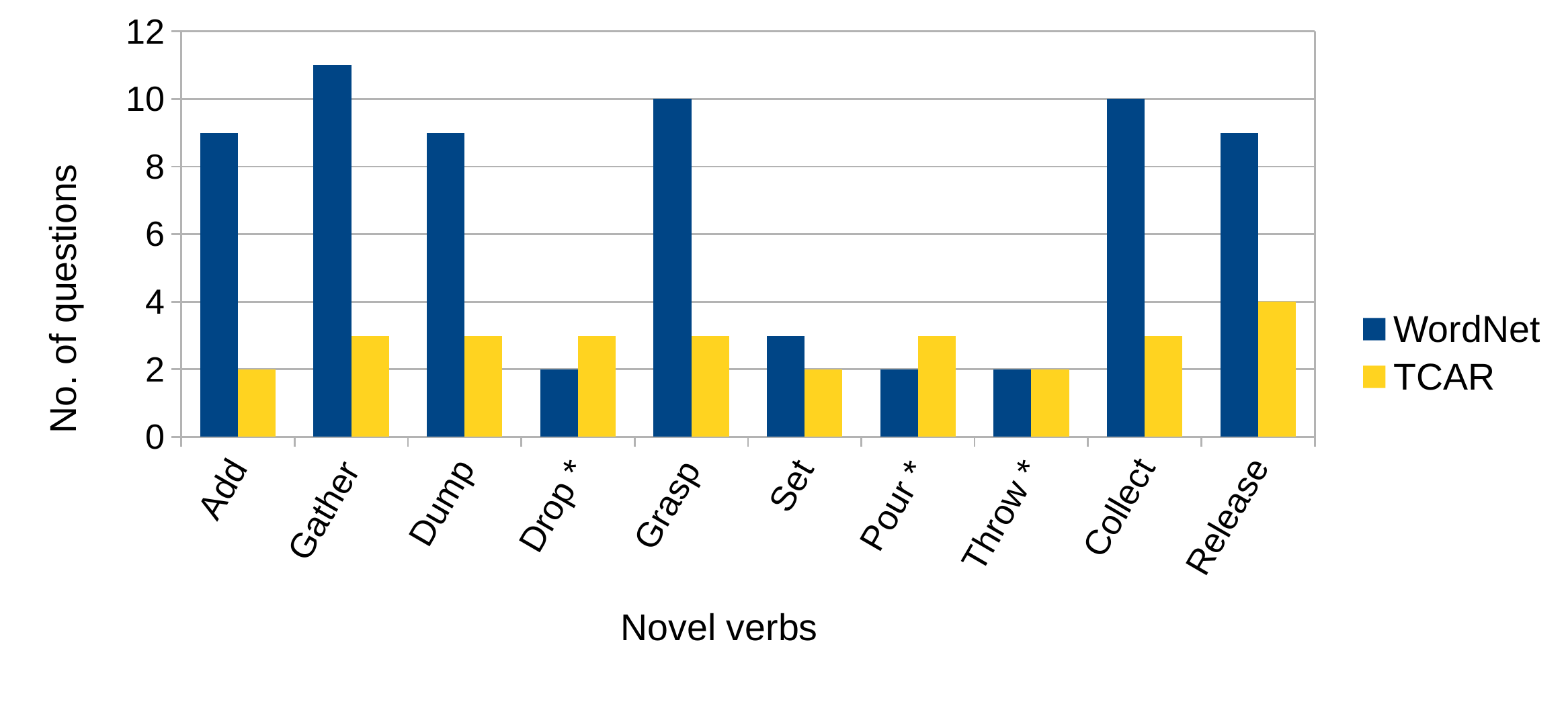}
	\caption{No of questions asked for instruction containing a novel verb. Verbs marked with * are synonymous with the (known set of) tasks.}
	\label{fig:novel-verbs}
\end{figure}

\begin{figure}[!t]
	\centering
	\includegraphics[width=\linewidth]{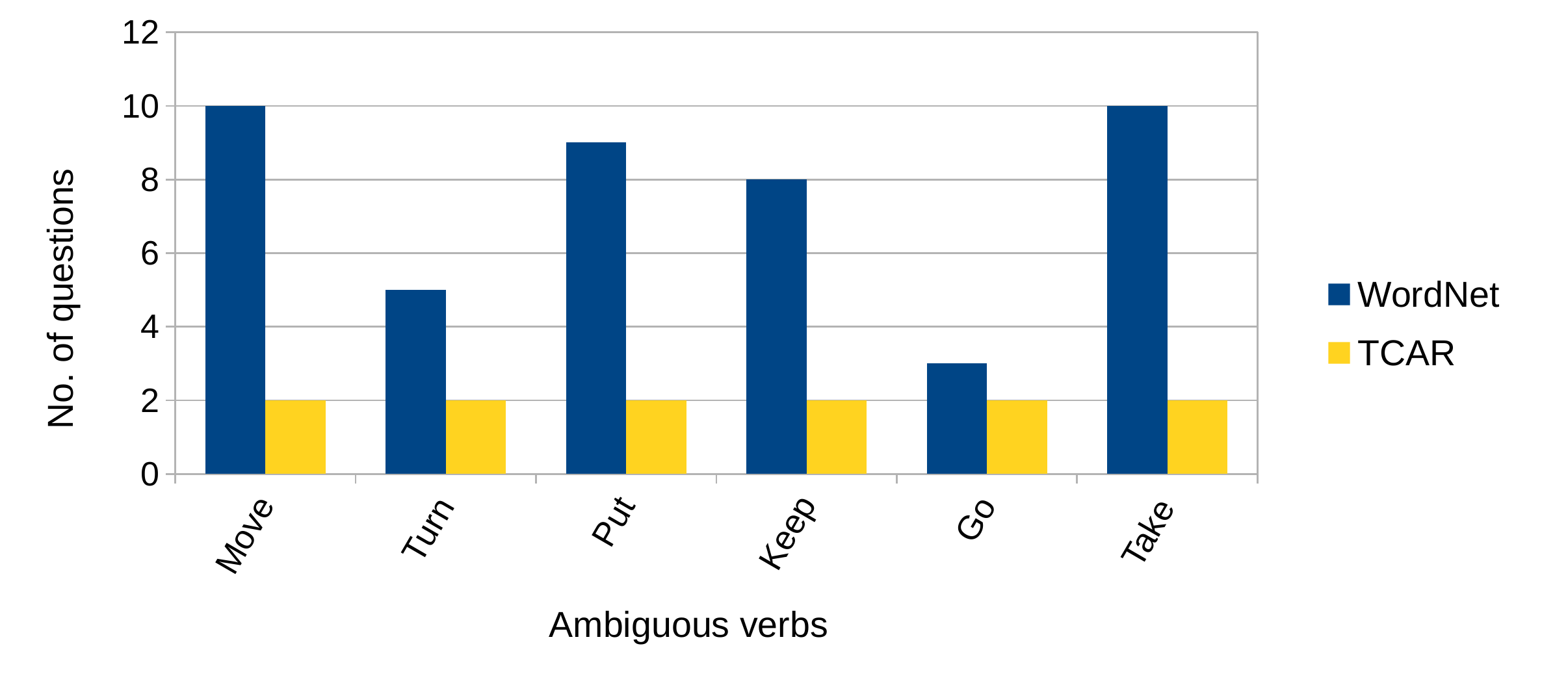}
	\caption{Number of questions asked for ambiguous instruction, which is initially misunderstood.}
	\label{fig:ambigous-verbs}
\end{figure}

We have evaluated the baseline and \system provided dialogue strategy using instructions from the VEIL dataset. The instructions containing novel verbs and their most similar task types are shown in Table~\ref{tab:novel-verbs}. For the ambiguous instructions, the original prediction by the task identification module and the actual task type retrieved through the dialogue strategy are shown in Table~\ref{tab:ambigous-verbs}. As shown in Fig.~\ref{fig:novel-verbs}, the WordNet baseline strategy asks 67 questions to understand novel verbs, whereas the strategy provided by \system asks only 27 questions for 10 instruction given in total. The WordNet baseline asks a smaller or similar set of questions when the novel verb is a synonym of the most common verb of the task. In cases where the novel verb is not a synonym but has a similar meaning in the context, \system always asks a much lesser number of questions. For ambiguous instructions, \system always asks a much lesser number of questions than the WordNet baseline, as shown in Fig.~\ref{fig:ambigous-verbs}. This is because \system exploits the task-argument relationships present in the training data to suggest the most likely alternatives. The WordNet baseline model asks 45 questions in total for the 6 ambiguous instructions shown in Table~\ref{tab:ambigous-verbs}, whereas \system asks only 12 questions.

It is important to note that the WordNet baseline model is unable to provide an answer when a non-expert user does not understand the question and asks for clarification. This is because the model calculates the task similarity score but can not predict the possible argument types that are required by the templates to correctly explain the task.

\section{CONCLUSIONS}
In this article, we describe our \textit{task conversational agent for robots (\system)} that performs task planning from natural language instructions. Specifically, we have presented a novel method to integrate a natural language parser and a dialogue agent with an automated planner. To enable such integration, we have developed a language understanding model to identify the intended tasks and the relevant parameters. We have presented a dialogue strategy to engage with a human participant when novel and/or ambiguous instructions are given. Our results suggest that our system benefited from context-aware reasoning of the knowledge base and dialogue to generate plans. Furthermore, our conversational agent engaged in conversations to understand instructions by asking for minimal and meaningful questions. Our system can further be extended to include probabilistic and hierarchical planning and also reasoning over a richer knowledge-base. One limitation of our dialogue strategy for task disambiguation is that it allows a single task intent in a novel instruction. We plan to extend this strategy for multiple task intents. There is also a scope of adopting this strategy to resolve planning and execution failures that require human-robot interaction.

\balance
\bibliographystyle{IEEEtran}
\bibliography{main}

\begin{thebibliography}{10}
\providecommand{\url}[1]{#1}
\csname url@samestyle\endcsname
\providecommand{\newblock}{\relax}
\providecommand{\bibinfo}[2]{#2}
\providecommand{\BIBentrySTDinterwordspacing}{\spaceskip=0pt\relax}
\providecommand{\BIBentryALTinterwordstretchfactor}{4}
\providecommand{\BIBentryALTinterwordspacing}{\spaceskip=\fontdimen2\font plus
\BIBentryALTinterwordstretchfactor\fontdimen3\font minus
  \fontdimen4\font\relax}
\providecommand{\BIBforeignlanguage}[2]{{%
\expandafter\ifx\csname l@#1\endcsname\relax
\typeout{** WARNING: IEEEtran.bst: No hyphenation pattern has been}%
\typeout{** loaded for the language `#1'. Using the pattern for}%
\typeout{** the default language instead.}%
\else
\language=\csname l@#1\endcsname
\fi
#2}}
\providecommand{\BIBdecl}{\relax}
\BIBdecl

\bibitem{pramanick2018defatigue}
P.~Pramanick and C.~Sarkar, ``Defatigue: Online non-intrusive fatigue detection
  by a robot co-worker,'' in \emph{2018 27th IEEE International Symposium on
  Robot and Human Interactive Communication (RO-MAN)}.\hskip 1em plus 0.5em
  minus 0.4em\relax IEEE, 2018, pp. 1129--1136.

\bibitem{vu2015companion}
C.~Vu, M.~Cross, T.~Bickmore, A.~Gruber, and T.~L. Campbell, ``Companion robot
  for personal interaction,'' 2015, uS Patent 8,935,006.

\bibitem{hinds2004whose}
P.~J. Hinds, T.~L. Roberts, and H.~Jones, ``Whose job is it anyway? a study of
  human-robot interaction in a collaborative task,'' \emph{Human-Computer
  Interaction}, vol.~19, no.~1, pp. 151--181, 2004.

\bibitem{matuszek2013learning}
C.~Matuszek, E.~Herbst, L.~Zettlemoyer, and D.~Fox, ``Learning to parse natural
  language commands to a robot control system,'' in \emph{Experimental
  Robotics}.\hskip 1em plus 0.5em minus 0.4em\relax Springer, 2013, pp.
  403--415.

\bibitem{thomason2015learning}
J.~Thomason, S.~Zhang, R.~Mooney, and P.~Stone, ``Learning to interpret natural
  language commands through human-robot dialog,'' in \emph{Proceedings of the
  24th International Conference on Artificial Intelligence}.\hskip 1em plus
  0.5em minus 0.4em\relax AAAI Press, 2015, pp. 1923--1929.

\bibitem{agarwal2017continuous}
S.~Agarwal, S.~Atreja, and G.~Dasgupta, ``Continuous learning as a service for
  conversational virtual agents,'' in \emph{International Conference on
  Service-Oriented Computing}.\hskip 1em plus 0.5em minus 0.4em\relax Springer,
  2017, pp. 641--656.

\bibitem{lam2017social}
C.~Lam and M.~A. Hannah, ``The social help desk: Examining how twitter is used
  as a technical support tool,'' \emph{Communication Design Quarterly Review},
  vol.~4, no.~2, pp. 37--51, 2017.

\bibitem{patidar2018automatic}
M.~Patidar, P.~Agarwal, L.~Vig, and G.~Shroff, ``Automatic conversational
  helpdesk solution using seq2seq and slot-filling models,'' in
  \emph{Proceedings of the 27th ACM International Conference on Information and
  Knowledge Management}.\hskip 1em plus 0.5em minus 0.4em\relax ACM, 2018, pp.
  1967--1975.

\bibitem{radlinski2017theoretical}
F.~Radlinski and N.~Craswell, ``A theoretical framework for conversational
  search,'' in \emph{Proceedings of the 2017 Conference on Conference Human
  Information Interaction and Retrieval}.\hskip 1em plus 0.5em minus
  0.4em\relax ACM, 2017, pp. 117--126.

\bibitem{paul2018temporal}
R.~Paul, A.~Barbu, S.~Felshin, B.~Katz, and N.~Roy, ``Temporal grounding graphs
  for language understanding with accrued visual-linguistic context,'' in
  \emph{Proceedings of the 26th International Joint Conference on Artificial
  Intelligence}, ser. IJCAI'17.\hskip 1em plus 0.5em minus 0.4em\relax AAAI
  Press, 2017, pp. 4506--4514.

\bibitem{lauria2002converting}
S.~Lauria, G.~Bugmann, T.~Kyriacou, J.~Bos, and E.~Klein, ``Converting natural
  language route instructions into robot executable procedures,'' in
  \emph{Robot and Human Interactive Communication, 2002. Proceedings. 11th IEEE
  International Workshop on}.\hskip 1em plus 0.5em minus 0.4em\relax IEEE,
  2002, pp. 223--228.

\bibitem{kollar2010toward}
T.~Kollar, S.~Tellex, D.~Roy, and N.~Roy, ``Toward understanding natural
  language directions,'' in \emph{Proceedings of the 5th ACM/IEEE international
  conference on Human-robot interaction}.\hskip 1em plus 0.5em minus
  0.4em\relax IEEE Press, 2010, pp. 259--266.

\bibitem{chen2011learning}
D.~L. Chen and R.~J. Mooney, ``Learning to interpret natural language
  navigation instructions from observations.'' in \emph{AAAI}, vol.~2, 2011,
  pp. 1--2.

\bibitem{baker1998berkeley}
C.~F. Baker, C.~J. Fillmore, and J.~B. Lowe, ``The berkeley framenet project,''
  in \emph{Proceedings of the 17th international conference on Computational
  linguistics-Volume 1}.\hskip 1em plus 0.5em minus 0.4em\relax Association for
  Computational Linguistics, 1998, pp. 86--90.

\bibitem{palmer2005proposition}
M.~Palmer, D.~Gildea, and P.~Kingsbury, ``The proposition bank: An annotated
  corpus of semantic roles,'' \emph{Computational linguistics}, vol.~31, no.~1,
  pp. 71--106, 2005.

\bibitem{scheutz2007first}
M.~Scheutz, P.~Schermerhorn, J.~Kramer, and D.~Anderson, ``First steps toward
  natural human-like hri,'' \emph{Autonomous Robots}, vol.~22, no.~4, pp.
  411--423, 2007.

\bibitem{thomas2012roboframenet}
B.~J. Thomas and O.~C. Jenkins, ``Roboframenet: Verb-centric semantics for
  actions in robot middleware,'' in \emph{Robotics and Automation (ICRA), 2012
  IEEE International Conference on}.\hskip 1em plus 0.5em minus 0.4em\relax
  IEEE, 2012, pp. 4750--4755.

\bibitem{bastianelli2016discriminative}
E.~Bastianelli, D.~Croce, A.~Vanzo, R.~Basili, and D.~Nardi, ``A discriminative
  approach to grounded spoken language understanding in interactive robotics,''
  in \emph{Proceedings of the Twenty-Fifth International Joint Conference on
  Artificial Intelligence}, ser. IJCAI'16.\hskip 1em plus 0.5em minus
  0.4em\relax AAAI Press, 2016, pp. 2747--2753.

\bibitem{bollini2013interpreting}
M.~Bollini, S.~Tellex, T.~Thompson, N.~Roy, and D.~Rus, ``Interpreting and
  executing recipes with a cooking robot,'' in \emph{Experimental
  Robotics}.\hskip 1em plus 0.5em minus 0.4em\relax Springer, 2013, pp.
  481--495.

\bibitem{antunes2016human}
A.~Antunes, L.~Jamone, G.~Saponaro, A.~Bernardino, and R.~Ventura, ``From human
  instructions to robot actions: Formulation of goals, affordances and
  probabilistic planning,'' in \emph{Robotics and Automation (ICRA), 2016 IEEE
  International Conference on}.\hskip 1em plus 0.5em minus 0.4em\relax IEEE,
  2016, pp. 5449--5454.

\bibitem{misra2016tell}
D.~K. Misra, J.~Sung, K.~Lee, and A.~Saxena, ``Tell me dave: Context-sensitive
  grounding of natural language to manipulation instructions,'' \emph{The
  International Journal of Robotics Research}, vol.~35, no. 1-3, pp. 281--300,
  2016.

\bibitem{chen2013handling}
X.-P. Chen, J.-M. Ji, Z.-Q. Sui, and J.-k. Xie, ``Handling open knowledge for
  service robots,'' in \emph{Twenty-Third International Joint Conference on
  Artificial Intelligence}, 2013.

\bibitem{lu2017interpreting}
D.~Lu and X.~Chen, ``Interpreting and extracting open knowledge for human-robot
  interaction,'' \emph{IEEE/CAA Journal of Automatica Sinica}, vol.~4, no.~4,
  pp. 686--695, 2017.

\bibitem{misra2015environment}
D.~K. Misra, K.~Tao, P.~Liang, and A.~Saxena, ``Environment-driven lexicon
  induction for high-level instructions,'' in \emph{Proceedings of the 53rd
  Annual Meeting of the Association for Computational Linguistics and the 7th
  International Joint Conference on Natural Language Processing (Volume 1: Long
  Papers)}, vol.~1, 2015, pp. 992--1002.

\bibitem{lu2017integrating}
D.~Lu, Y.~Zhou, F.~Wu, Z.~Zhang, and X.~Chen, ``Integrating answer set
  programming with semantic dictionaries for robot task planning,'' in
  \emph{Proceedings of the 26th International Joint Conference on Artificial
  Intelligence}.\hskip 1em plus 0.5em minus 0.4em\relax AAAI Press, 2017, pp.
  4361--4367.

\bibitem{Munawar2018}
A.~Munawar, G.~D. Magistris, T.~Pham, D.~Kimura, M.~Tatsubori, T.~Moriyama,
  R.~Tachibana, and G.~Booch, ``Maestrob: A robotics framework for integrated
  orchestration of low-level control and high-level reasoning,'' in \emph{2018
  IEEE International Conference on Robotics and Automation (ICRA)}, May 2018,
  pp. 527--534.

\bibitem{mcdermott1998pddl}
M.~Ghallab, A.~Howe, C.~Knoblock, D.~McDermott, A.~Ram, M.~Veloso, D.~Weld, and
  D.~Wilkins, ``Pddl-the planning domain definition language,'' \emph{AIPS-98
  planning committee}, vol.~3, p.~14, 1998.

\bibitem{kollar2013learning}
T.~Kollar, V.~Perera, D.~Nardi, and M.~Veloso, ``Learning environmental
  knowledge from task-based human-robot dialog,'' in \emph{Robotics and
  Automation (ICRA), 2013 IEEE International Conference on}.\hskip 1em plus
  0.5em minus 0.4em\relax IEEE, 2013, pp. 4304--4309.

\bibitem{liu2018generating}
R.~Liu and X.~Zhang, ``Generating machine-executable plans from end-user's
  natural-language instructions,'' \emph{Knowledge-Based Systems}, vol. 140,
  pp. 15--26, 2018.

\bibitem{hoffmann2001ff}
J.~Hoffmann and B.~Nebel, ``The ff planning system: Fast plan generation
  through heuristic search,'' \emph{Journal of Artificial Intelligence
  Research}, vol.~14, pp. 253--302, 2001.

\bibitem{bastianelli2014huric}
E.~Bastianelli, G.~Castellucci, D.~Croce, L.~Iocchi, R.~Basili, and D.~Nardi,
  ``Huric: a human robot interaction corpus,'' in \emph{Proceedings of the
  Ninth International Conference on Language Resources and Evaluation
  (LREC-2014)}, 2014.

\bibitem{miller1995wordnet}
G.~A. Miller, ``Wordnet: a lexical database for english,'' \emph{Communications
  of the ACM}, vol.~38, no.~11, pp. 39--41, 1995.

\end{thebibliography}

\end{document}